\documentclass[conference]{IEEEtran}
\IEEEoverridecommandlockouts
\usepackage{float}
\usepackage{array} 
\usepackage{cite}
\usepackage{amsmath,amssymb,amsfonts}
\usepackage{algorithmic}
\usepackage{graphicx}
\usepackage{textcomp}
\usepackage{xcolor}
\usepackage{url}
\usepackage[utf8]{inputenc}
\usepackage{tabularx} 
\usepackage{booktabs} 
\def\BibTeX{{\rm B\kern-.05em{\sc i\kern-.025em b}\kern-.08em
    T\kern-.1667em\lower.7ex\hbox{E}\kern-.125emX}}
\begin{document}

\title{Optimizing Large Language Models for Turkish: New Methodologies in Corpus Selection and Training
}

\author{
\IEEEauthorblockN{H. Toprak Kesgin}
\IEEEauthorblockA{\textit{Department of Computer Engineering} \\
\textit{Yildiz Technical University}\\
Istanbul, Turkey
 \\
tkesgin@yildiz.edu.tr}
\and
\IEEEauthorblockN{M. Kaan Yuce}
\IEEEauthorblockA{\textit{Department of Computer Engineering} \\
\textit{Yildiz Technical University}\\
Istanbul, Turkey \\
kaan.yuce@yildiz.edu.tr}
\and
\IEEEauthorblockN{Eren Dogan}
\IEEEauthorblockA{\textit{Department of Computer Engineering} \\
\textit{Yildiz Technical University}\\
Istanbul, Turkey \\
eren.dogan2@std.yildiz.edu.tr}
\and
\IEEEauthorblockN{M. Egemen Uzun}
\IEEEauthorblockA{\textit{Department of Computer Engineering} \\
\textit{Yildiz Technical University}\\
Istanbul, Turkey \\
egemen.uzun@std.yildiz.edu.tr}
\and
\IEEEauthorblockN{Atahan Uz}
\IEEEauthorblockA{\textit{Department of Computer Engineering} \\
\textit{Yildiz Technical University}\\
Istanbul, Turkey\\
atahan.uz@std.yildiz.edu.tr}
\and
\IEEEauthorblockN{Elif İnce}
\IEEEauthorblockA{\textit{Department of Computer Engineering} \\
\textit{Yildiz Technical University}\\
Istanbul, Turkey \\
elif.ince@std.yildiz.edu.tr}
\and
\IEEEauthorblockN{Yusuf Erdem}
\IEEEauthorblockA{\textit{Department of Computer Engineering} \\
\textit{Yildiz Technical University}\\
Istanbul, Turkey \\
erdem.erdem@std.yildiz.edu.tr}
\and
\IEEEauthorblockN{Osama Shbib}
\IEEEauthorblockA{\textit{Department of Computer Engineering} \\
\textit{Yildiz Technical University}\\
Istanbul, Turkey \\
osama.shbib@std.yildiz.edu.tr}
\and
\IEEEauthorblockN{Ahmed Zeer}
\IEEEauthorblockA{\textit{Department of Computer Engineering} \\
\textit{Yildiz Technical University}\\
Istanbul, Turkey \\
ahmed.zeer@std.yildiz.edu.tr}
\and
\IEEEauthorblockN{M. Fatih Amasyali}
\IEEEauthorblockA{\textit{Department of Computer Engineering} \\
\textit{Yildiz Technical University}\\
Istanbul, Turkey \\
amasyali@yildiz.edu.tr}
}

\IEEEoverridecommandlockouts \IEEEpubid{\makebox[\columnwidth]{979-8-3315-3149-2/24/\$31.00~\copyright2024 IEEE\hfill} \hspace{\columnsep}\makebox[\columnwidth]{ }}
\maketitle

\begin{abstract}
In this study, we develop and assess new corpus selection and training methodologies to improve the effectiveness of Turkish language models. Specifically, we adapted Large Language Model generated datasets and translated English datasets into Turkish, integrating these resources into the training process. This approach led to substantial enhancements in model accuracy for both few-shot and zero-shot learning scenarios. Furthermore, the merging of these adapted models was found to markedly improve their performance. Human evaluative metrics, including task-specific performance assessments, further demonstrated that these adapted models possess a greater aptitude for comprehending the Turkish language and addressing logic-based queries. This research underscores the importance of refining corpus selection strategies to optimize the performance of multilingual models, particularly for under-resourced languages like Turkish.
\end{abstract}

\begin{IEEEkeywords}
Natural Language Processing, Multilingual Models, Large Language Model Optimization, Turkish Language Models, Cross-Lingual Transfer Learning, Few-Shot Learning, Zero-Shot Learning, Synthetic Datasets
\end{IEEEkeywords}

\section{Introduction}
In recent years, the development of language models has been an important area of research in artificial intelligence. In particular, multilingual models have the potential to overcome data gaps in different languages and improve language learning processes\cite{qin2024multilingual}. To realize this potential, it is important to evaluate and optimize the performance of multilingual models.
Multilingual models leverage shared knowledge across languages, allowing for more robust and comprehensive language understanding \cite{zhao2024llama}. However, for these models to be effective, there must be sufficient quantity and quality of data in each language. There can be large differences between the quantity and quality of data in different languages, which can cause models to underperform in some languages. For languages with limited data sources, such as Turkish, multilingual models must be trained and optimized to overcome these shortcomings. The focus of this study is to improve the performance of Turkish language in multilingual models and enable them to produce more accurate responses. In this context, we investigate how we can improve the performance of existing multilingual models for Turkish. 


In this study, English datasets, which were found to enhance the performance of the models, were translated into Turkish and utilized in the training of the models. After training, the performance of the models was evaluated both by human evaluation and specifically on Turkish adapted datasets with few-shot datasets such as HellaSwag \cite{zellers2019hellaswag} and ARC\cite{clark2018think}. In particular, Accuracy metric were used as corpus selection, model training and evaluation criteria. The results of the study demonstrate the observed improvements in the performance of Turkish language models. Furthermore, illustrates the positive effects of large-scale models on small-scale language models, as well as the benefits of synthetic and translation datasets. In conclusion, this study makes significant contributions to corpus selection methodologies and training strategies for the development of Turkish language models, with the aim of increasing the effectiveness of Turkish in the field of language technologies.

The main contributions presented in this paper are as follows:
\begin{itemize}

\item By enhancing existing corpus selection methodologies and adapting them to Turkish, we have devised novel, optimized methods for language modeling.

\item New adaptation datasets in Turkish were also created. New datasets, meticulously translated and harmonized from English to Turkish, were created for the purpose of training Turkish language models.

\item New pre-training corpora have been designed for Turkish language models, with the objective of improving model performance.

\item A comprehensive comparison has been conducted between existing models and models trained with the proposed method, in addition to other Turkish language models. This comparison has been conducted using both human voting and evaluations based on few-shot approaches.
\end{itemize}

\section{Corpus Creation}
Zero-shot and few-shot methods are commonly employed to assess the performance of large language models.
In this study, we concentrated on enhancing the performance of Turkish models on these datasets.
In our initial study, we sought models that are relatively small in size but demonstrate relative proficiency on few-shot evaluation datasets.
Model selection was primarily based on the Cosmopedia dataset \cite{benallal2024cosmopedia}, which also served as the main training dataset for the Cosmo1b model.
The Cosmopedia dataset comprises approximately 30 million files and 25 billion tokens.
It was created using the Mixtral-8x7B-Instruct-v0.1 model \cite{jiang2023mistral} and comprises eight subsets of varying sizes, derived from sources including synthetic textbooks, blog posts, stories, posts, and WikiHow articles.
The number of examples and the size of the text in each subset are provided in Table \ref{tab:dataset_details}.
Furthermore, the OpenOrca dataset \cite{mukherjee2023orca}, an open-source instruction completion dataset comprising 3.7GB of text, was employed for the model to execute human instructions.
All these datasets are in English and translated into Turkish using the Google Translate API.

\begin{table}[ht]
\centering
\caption{Number of examples and size of text in each subset of the Cosmopedia dataset}
\label{tab:dataset_details}
\begin{tabular}{|l|r|r|}
\hline
\textbf{Subset}          & \textbf{Sample} & \textbf{Size}  \\ \hline
auto\_math\_text         & 1.95M           & 8.8GB          \\ \hline
khanacademy              & 24.1K           & 125MB          \\ \hline
openstax                 & 126K            & 700MB          \\ \hline
stanford                 & 1.02M           & 6.6GB          \\ \hline
stories                  & 4.99M           & 21.8GB         \\ \hline
web\_samples\_v1         & 12.4M           & 71.9GB         \\ \hline
web\_samples\_v2         & 10.3M           & 30.5GB         \\ \hline
wikihow                  & 179K            & 1GB            \\ \hline
\end{tabular}
\end{table}

Given their relatively small sizes, the Stanford, Khan Academy, WikiHow, and OpenStax datasets were combined to achieve meaningful results. This combination of datasets was used in our evaluation process as detailed in Table \ref{tab:performance_comparison}. 'SKWO' stands for the combined datasets of Stanford, Khan Academy, WikiHow, and OpenStax.

Training models with parameters as large as 7 billion can present a significant challenge, due to the limitations of both hardware and time.
To mitigate these challenges, we employed smaller yet effective versions of the models for initial testing phases.
Accordingly, in the present study, we employed the following methodology to ascertain which types of datasets enhance the few-shot performance of LLMs.
A model was trained ten times smaller with the specified dataset, and the performance of the model on few-shot datasets was compared before and after training.
It is hypothesized that if a dataset can enhance the performance of a smaller model, it can also enhance the performance of a larger model.
In all experiments, the ytu-ce-cosmos/turkish-gpt2-large model \cite{kesgin2024introducing}, trained exclusively for the Turkish language with 750 million parameters, was utilized.
The datasets selected for evaluation were chosen based on three criteria: (1) they are widely used, (2) they are datasets that will not lose their meaning when translated, and (3) the ytu-ce-cosmos/turkish-gpt2-large model used can perform better than random prediction.
These criteria ensured that the selected datasets were both relevant and suitable for assessing the performance of the model when adapted to Turkish.
The model demonstrated comparable performance to random prediction in certain datasets.
Consequently, the datasets selected for evaluating the model were COPA\cite{ponti2020xcopa}, XStoryCloze\cite{lin2022few}, ARC Easy, ARC \cite{clark2018think}, and HellaSwag \cite{zellers2019hellaswag}. 

\begin{table}[ht]
\centering
\caption{Evaluation of Model Performance on Different Datasets}
\label{tab:performance_comparison}
\resizebox{\columnwidth}{!}{%
  \begin{tabular}{|l|r|r|r|r|r|r|}
  \hline
 \textbf{Data} & \textbf{Copa} & \textbf{Xstory} &\textbf{ARC Easy} & \textbf{ARC} & \textbf{HellaSwag} & \textbf{Avg} \\ \hline
Base & 60.00 & 55.33 & 37.71 & 23.65 & 36.39 & 42.62 \\ \hline
SKWO & 59.20 & 57.64 & 39.31 & 27.92 & 36.19 & 44.05 \\ \hline
AutoMath & 57.20 & 53.47 & 36.06 & 27.50 & 33.62 & 41.57 \\ \hline
Stories & 59.40 & 60.95 & 42.14 & 25.70 & 37.80 & 45.20 \\ \hline
Web1 & 57.00 & 55.85 & 38.04 & 24.34 & 36.36 & 42.32 \\ \hline
Web2 & 58.00 & 56.32 & 39.10 & 24.77 & 36.57 & 42.95 \\ \hline
OpenOrca & 59.40 & 56.05 & 38.47 & 24.77 & 37.22 & 43.18 \\ \hline
  \end{tabular}
}
\end{table}

To ensure a fair and comprehensive evaluation, ARC dataset was used as 25 shots, HellaSwag as 5 shots, and all other datasets as zero shots. We then evaluated the data set as follows. After training, we averaged the 5 accuracy scores. These scores can be seen in Table \ref{tab:performance_comparison}. If the base model improved the success of the model, we marked this dataset as good and selected it. The selected data sets are SKWO, Stories, OpenOrca. 
This selection highlights datasets that not only challenge the model but also contribute significantly to its ability to understand and respond to complex instructions in Turkish. These findings underscore the importance of tailored dataset composition in enhancing language model performance for specific linguistic tasks.

\section{Training with Corpus Selection}
We selected the Llama3-8b model \cite{llama3modelcard}, which gives the highest few-shot scores for Turkish, and the instruct versions of the same model to test whether the models that increase success in this 750m parameter model can also increase success in larger models.
We trained this model using the fullfine training with the SKWO, Stories and OpenOrca data sets selected according to Table \ref{tab:dataset_details}.
During training, we only tracked the performance on the ARC dataset due to resource and time constraints.
We saw that the success of the model gradually increased with the checkpoints we received during training.
As a result, our base model, with an accuracy of 48.72\% on the instruction dataset, achieved 49.15\% on the ARC dataset.
Best of our knowledge, our Base and Our Instruct models are the highest performing open-source Turkish models in the range of 7-8 billion parameters. 

Training details are as follows.
Training was conducted over 1 epoch to optimize learning within the constraints of our resources and timelines.
The decision to use a batch size of 1 with gradient accumulation set at 512 was guided by our hardware limitations and the need to handle large-scale data efficiently.
A learning rate of $1e-6$ was chosen based on preliminary tests that indicated it balances training speed and model stability effectively.
Gradient clipping was set at $0.05$ to prevent the exploding gradient problem, enhancing the stability of training over extensive datasets.
The 8-bit AdamW \cite{loshchilov2017decoupled} optimizer was selected for its efficiency in handling large models and datasets, providing a balance between computational demand and performance.

\section{Model Merging}

Model merging is an important technique that has emerged recently \cite{goddard2024arcee}.
In recent advancements, model merging has proven effective in enhancing model performance by combining the strengths of various trained models. 
In this technique, the weights of models trained with different data sets with the same architecture can be combined with different techniques.
After these combinations, the combined model can be more successful than the 2 models.
For this purpose, this technique was employed to merge our trained models with the base and instruct versions of Llama3, resulting in significant performance improvements.
The linear merging method, which is the classic merge method, was used for this purpose.
The few-shot and human voting comparisons of the two fine-tuned models and their merged versions are described in detail in \textit{Comparison of Turkish Language Models} section.
This comprehensive evaluation aims to provide empirical evidence on the effectiveness of model merging as a viable method for enhancing language model performance.

\section{Comparison of Turkish language models}
In this section, we present a detailed comparison of Turkish language models to evaluate the effectiveness of our proposed methods. The comparison is carried out using both dataset performance metrics and human judge evaluations to provide a comprehensive understanding of the improvements achieved. The models compared include existing Turkish language models, models trained using traditional methods, and the models developed through our proposed approach.

\subsection{Human Judge Evaluation Metrics}

We employed metrics like the ELO score and Win Percentage in human assessments to gauge model performance. ELO scores are used to determine the comparative skill levels of participants in zero-sum games. The ELO metric is widely used to compare the performance of language models against each other based on human voting\cite{chiang2024chatbot}. In this process, judges assessed responses from the models to questions in the $\mathbf{V}$ dataset.

\begin{table}[ht]
\caption{Random Samples from V Dataset}
\centering
\scriptsize 
\begin{tabular}{|>{\centering\arraybackslash}m{2cm}|>{\centering\arraybackslash}m{5cm}|}
\hline

\textbf{Category} & \textbf{Instruction} \\
\hline

Sentiment Analysis & Describe a student's emotions when receiving an acceptance letter. \\
\noalign{\global\arrayrulewidth=0pt}
\cline{2-2}
\noalign{\global\arrayrulewidth=.4pt}
\textit{Duygu Analizi} & \textit{Bir öğrencinin kabul mektubu aldığında hissettiği duyguları tasvir ediniz.} \\
\hline

Coding & Write a C code to check if a word is a palindrome. \\
\noalign{\global\arrayrulewidth=0pt}
\cline{2-2}
\noalign{\global\arrayrulewidth=.4pt}
\textit{Kod Yazma} & \textit{Bir kelimenin palindrome olup olmadığını kontrol eden bir C kodu yazınız.} \\
\hline

Style and Tone Change & Rewrite "I went to the park last week." in future tense. \\
\noalign{\global\arrayrulewidth=0pt}
\cline{2-2}
\noalign{\global\arrayrulewidth=.4pt}
\textit{Stil ve Ton Değişikliği} & \textit{"Geçen hafta parka gittim." cümlesini gelecek zaman kullanarak yeniden yazın.} \\
\hline

Story Creation & Tell a disaster survivor's story of finding hope. \\
\noalign{\global\arrayrulewidth=0pt}
\cline{2-2}
\noalign{\global\arrayrulewidth=.4pt}
\textit{Hikaye Oluşturma} & \textit{Bir felaket sonrası hayatta kalan birinin umudu bulma çabasını anlatın.} \\
\hline

Sentence Completion & Complete: While searching for a treasure hidden in the sea, \\
\noalign{\global\arrayrulewidth=0pt}
\cline{2-2}
\noalign{\global\arrayrulewidth=.4pt}
\textit{Cümle Tamamlama} & \textit{Cümleyi tamamla: Denizin derinliklerinde saklı bir hazineyi ararken,} \\
\hline

Title Creation & Write a title for an article on technology's effects on children. \\
\noalign{\global\arrayrulewidth=0pt}
\cline{2-2}
\noalign{\global\arrayrulewidth=.4pt}
\textit{Başlık Oluşturma} & \textit{Teknolojinin çocuklar üzerindeki etkilerini tartışan bir makale için başlık yazın.} \\
\hline

Listing & List 5 interesting science project ideas for students. \\
\noalign{\global\arrayrulewidth=0pt}
\cline{2-2}
\noalign{\global\arrayrulewidth=.4pt}
\textit{Listeleme} & \textit{Öğrenciler için 5 ilginç bilim projesi fikri listeleyin.} \\
\hline

Basic Math & A device lasts 40 hours. How long will it last with 50\% charge? \\
\noalign{\global\arrayrulewidth=0pt}
\cline{2-2}
\noalign{\global\arrayrulewidth=.4pt}
\textit{Basit Matematik} & \textit{Bir cihaz 40 saat dayanıyor. \%50 şarjla ne kadar dayanır?} \\
\hline

Logic & If blue birds can't fly, what color can a flying bird be? \\
\noalign{\global\arrayrulewidth=0pt}
\cline{2-2}
\noalign{\global\arrayrulewidth=.4pt}
\textit{Mantık} & \textit{Mavi kuşlar uçamıyorsa, uçabilen bir kuşun rengi ne olabilir?} \\
\hline

Explaining & What steps are needed to write a novel? \\
\noalign{\global\arrayrulewidth=0pt}
\cline{2-2}
\noalign{\global\arrayrulewidth=.4pt}
\textit{Açıklama} & \textit{Bir roman yazmak için gerekli adımlar nelerdir?} \\
\hline

How to & What are efficient ways to take notes? \\
\noalign{\global\arrayrulewidth=0pt}
\cline{2-2}
\noalign{\global\arrayrulewidth=.4pt}
\textit{Nasıl Yapılır} & \textit{Verimli not tutmanın yolları nelerdir?} \\
\hline

Intermediate Math & How long will it take to fill a pool if one tap fills it in 3 hours and another tap empties it in 6 hours? \\
\noalign{\global\arrayrulewidth=0pt}
\cline{2-2}
\noalign{\global\arrayrulewidth=.4pt}
\textit{Orta Düzey Matematik} & \textit{Bir musluk havuzu 3 saatte doldurur, diğeri 6 saatte boşaltırsa, havuz ne kadar sürede dolar?} \\

\noalign{\global\arrayrulewidth=0pt}
\cline{2-2}
\noalign{\global\arrayrulewidth=.4pt}
\hline

\end{tabular}
\label{tab:V_dataset}
\end{table}

Each judge was presented with a random question and two different model responses, with the model names concealed to maintain impartiality. Initially, each model started with an ELO rating of 1000. When models were compared, the preferred model gained ELO points while the other lost points. Defeating a high ELO model grants more points than defeating a low ELO model. This system effectively illustrates the relative strengths of the models. A high ELO score signifies superior performance relative to other models. To accurately capture the ELO results, we randomly reordered the matchups in the dataset and recalculated the models’ ELO scores across 1000 different scenarios. 
Each permutation represented a scenario with completely random matchup orders. We then calculated the averages and confidence intervals for each model’s ELO scores across these scenarios, allowing us to understand the potential impact of matchup orders on ELO scores and to generalize the results more effectively. Eight judges participated equally in this evaluation, casting a total of $3000$ votes to ensure a thorough and balanced assessment.

Win Percentage (Winpct) measures a model’s success against other models based on human votes. This metric calculates the ratio of votes a model receives to the total number of votes:

\begin{center}
\normalsize{$\text{{winpct}} = \frac{{\text{{win}} + \text{{both}}}}{{\text{{total}}}}$}
\end{center}

These metrics and evaluation methods enable us to compare the performance of language models across different task and scenarios.

\subsection{Comparison on Few-Shot Datasets}

To assess the performance of the language models, we conducted evaluations on several datasets specifically adapted for Turkish. These datasets were meticulously translated and harmonized from their English counterparts to ensure consistency and relevance. The key datasets used for this comparison are HellaSwag\cite{zellers2019hellaswag}, ARC\cite{clark2018think}, GSM8K\cite{cobbe2021gsm8k}, MMLU\cite{hendryckstest2021}, Truthful\_qa\cite{lin2021truthfulqa} and Winogrande\cite{ai2:winogrande} which were adapted for few-shot learning scenarios.
The relevant model scores are shown in Table~\ref{evaluationDatasets}

Model Selection: The models selected for comparison include our finetuned versions of the Llama3 and Llama3-Instruct models Our Base and Our Instruct. The original Llama3 and Llama3-Instruct models, which serve as the baseline. Additionally, we included three of the most successful Turkish models as identified in \cite{dogan2024turkcce}: Trendyol Chat\cite{Trendyol}, Turkcell\cite{Turkcell}, and SambaLingo\cite{Sambalingo}. Our evaluation also incorporates both the Base and Instruct versions of our newly trained models, as well as their merged versions with the Llama3-Instruct model\cite{llama3modelcard}. This selection provides a comprehensive view of how well different approaches perform on the Turkish adapted datasets.

\begin{table}[ht]
\centering
\caption{Results of models on evaluation datasets}
\footnotesize 
\setlength{\tabcolsep}{4pt} 
\newcommand\myfontsize{\fontsize{7pt}{8.4pt}\selectfont}
\begin{tabular}{
  |>{\centering\arraybackslash}m{0.9cm}
  |>{\centering\arraybackslash}m{0.5cm}
  |>{\centering\arraybackslash}m{1.02cm}
  |>{\centering\arraybackslash}m{0.75cm}
  |>{\centering\arraybackslash}m{0.68cm}
  |>{\centering\arraybackslash}m{1.1cm}
  |>{\centering\arraybackslash}m{1.12cm}
  |>{\centering\arraybackslash}m{0.5cm}|
}
\hline

{\myfontsize\textbf{Model /Dataset}} & {\myfontsize\textbf{ARC}} & {\textbf{Hella
Swag }} & {\myfontsize\textbf{GSM8K}} & {\myfontsize\textbf{MMLU}} & {\myfontsize\textbf{Truthful\_qa}} & {\myfontsize\textbf{Winogrande}} & {\myfontsize\textbf{Avg}} \\ 
\hline
\myfontsize Llama3 Instruct & \myfontsize 44.20 & \myfontsize 44.90 & \myfontsize 54.29 & \myfontsize 50.91 & \myfontsize 50.43 & \myfontsize 50.43 & \myfontsize 50.05 \\ 
\hline
\myfontsize Our Base Model & \myfontsize 48.72 &\myfontsize 50.45 &  \myfontsize 48.44 &  \myfontsize 51.99 & \myfontsize 49.86 & \myfontsize 57.74 & \myfontsize 51.20 \\ 
\hline
\myfontsize Our Inst Model & \myfontsize 49.15 & \myfontsize 50.76 & \myfontsize55.43 & \myfontsize53.23 & \myfontsize48.89 & \myfontsize58.73 & \myfontsize52.70 \\ 
\hline
\myfontsize Our Merged Base Model & \myfontsize49.40 & \myfontsize 51.00 &\myfontsize 58.47 &\myfontsize 53.37 &\myfontsize 49.88 &\myfontsize 56.40 & \myfontsize 53.09 \\ 
\hline
\myfontsize Our Merged Inst Model &\myfontsize  48.98 & \myfontsize \myfontsize 50.45 & \myfontsize  57.10 & \myfontsize53.37 & \myfontsize49.88 & \myfontsize56.40 & \myfontsize 52.70 \\ 
\hline
\myfontsize \myfontsize Samba Lingo & \myfontsize 44.97 & \myfontsize 55.43 &\myfontsize 4.94 & \myfontsize36.40 & \myfontsize44.08 & \myfontsize58.14 &\myfontsize 40.66 \\ 
\hline
\myfontsize Trendyol Chat & \myfontsize34.04 & \myfontsize41.65 & \myfontsize1.97 & \myfontsize 34.01 & \myfontsize 42.20 & \myfontsize 54.34 & \myfontsize 34.70 \\ 
\hline
\myfontsize Turkcell & \myfontsize 43.43 &\myfontsize 49.19 &\myfontsize 23.84 &\myfontsize 40.90 &\myfontsize 41.62 & \myfontsize56.56 & \myfontsize42.59 \\ 
\hline
\end{tabular}
\hspace{3cm}
\label{evaluationDatasets}  
\end{table}

The Table~\ref{evaluationDatasets} illustrates that the models developed through our proposed methods significantly outperform both the existing Turkish models and those trained using traditional methods. This indicates the effectiveness of the new pre-training corpora, the optimized corpus selection and model merging methodologies.
\subsection{Comparison with Human Judge Voting}

\begin{figure*}[htbp]
    \centering
    \includegraphics[width=0.75\linewidth]{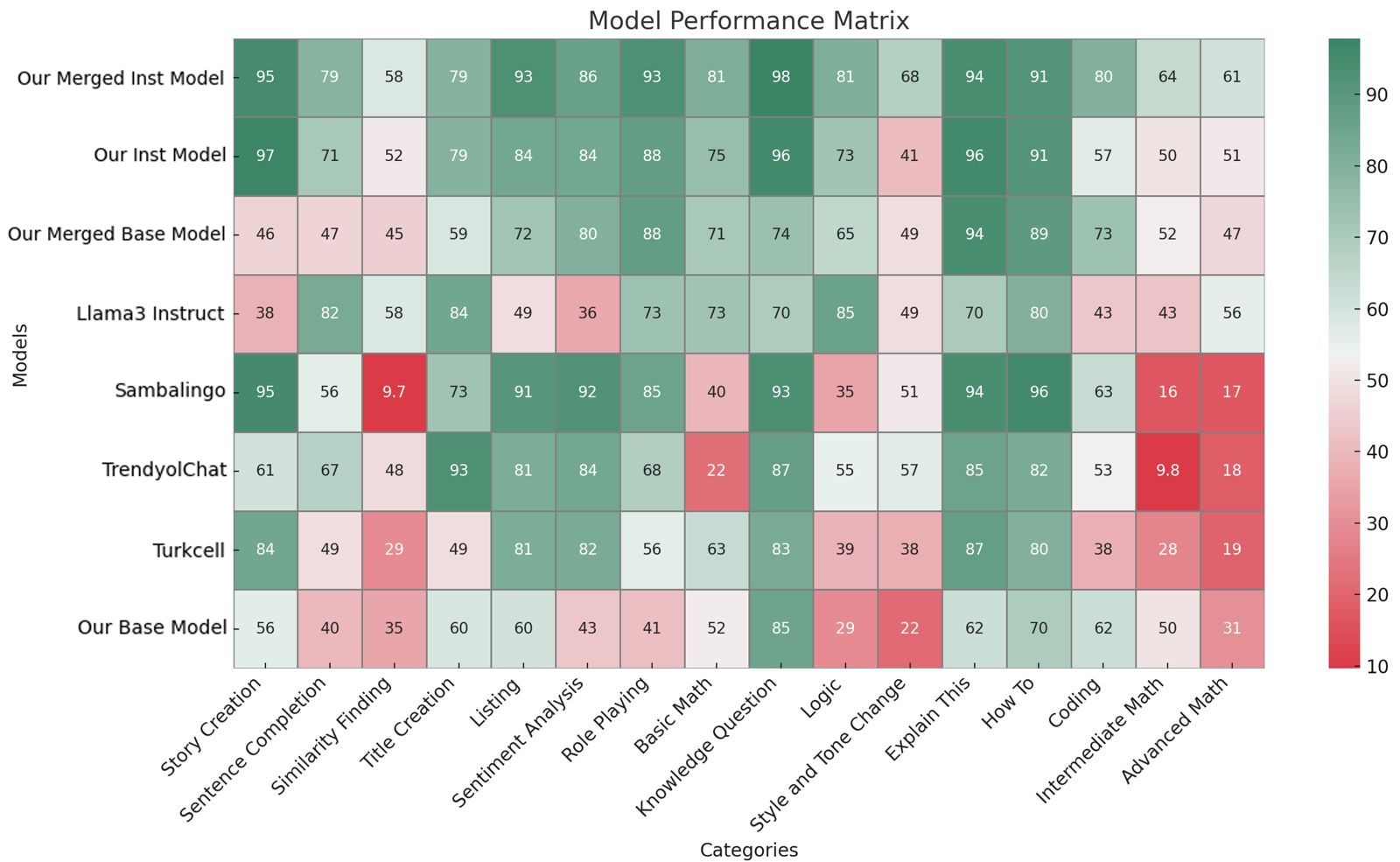}
    \caption{Models Performance Across Categories}
    \label{fig:categorical-performance}
\end{figure*}

In addition to quantitative metrics, we also conducted evaluations through human judge voting to obtain qualitative insights into the model's performance. Human judges evaluated the models based on their responses to various tasks, considering aspects such as creativity, math, logic, and finding similarities. The models' ELO ratings and winning percentages are shown in Table~\ref{ratings}. The scores for all tasks are shown in Figure ~\ref{fig:categorical-performance}, showing the superiority of the models over each other in different areas. 

\begin{table}[ht]
\centering
\caption{Results of the voting of models}
\begin{tabular}{|p{3.5cm}|p{1cm}|p{1.5cm}|p{1cm}|}
\hline
\textbf{Model}    & \textbf{ELO} & \textbf{Confidence Interval} & \textbf{WinPct} \\ \hline


Our Merged Inst-Model & 1061 & +61/-52 & 80.59 \\ \hline
Our Inst-Model & 1039 & +51/-55 & 73.56 \\ \hline
Our Merged Base-Model & 1004 & +53/-55 & 62.88 \\ \hline

Llama3 Instruct  & 995 & +57/-55 & 60.47 \\ \hline
Sambalingo  & 987 & +54/-54 & 60.47 \\ \hline
TrendyolChat & 983 & +56/-60 & 58.57 \\ \hline
Turkcell & 972 & +54/-58 & 55.25 \\ \hline
Our Base-Model  & 902 & +63/-59 & 48.8  \\ \hline
\end{tabular}
\hspace{3cm}
\label{ratings}
\end{table}

The results indicate that our models are highly favored by human evaluators compared to existing Turkish models and those trained with traditional techniques. This preference underscores the efficacy of our novel pre-training corpora, optimized corpus selection, and model merging methodology. As illustrated in Figure~\ref{fig:categorical-performance}, our models exhibit superior performance across various categories, including \textit{Similarity Finding}, \textit{Logic}, \textit{Intermediate Math}, and \textit{Advanced Math}, highlighting their exceptional problem-solving and reasoning capabilities.
\\

\subsection{Correlations}

\begin{figure}[htbp]
    \centering
    \includegraphics[width=0.75\linewidth]{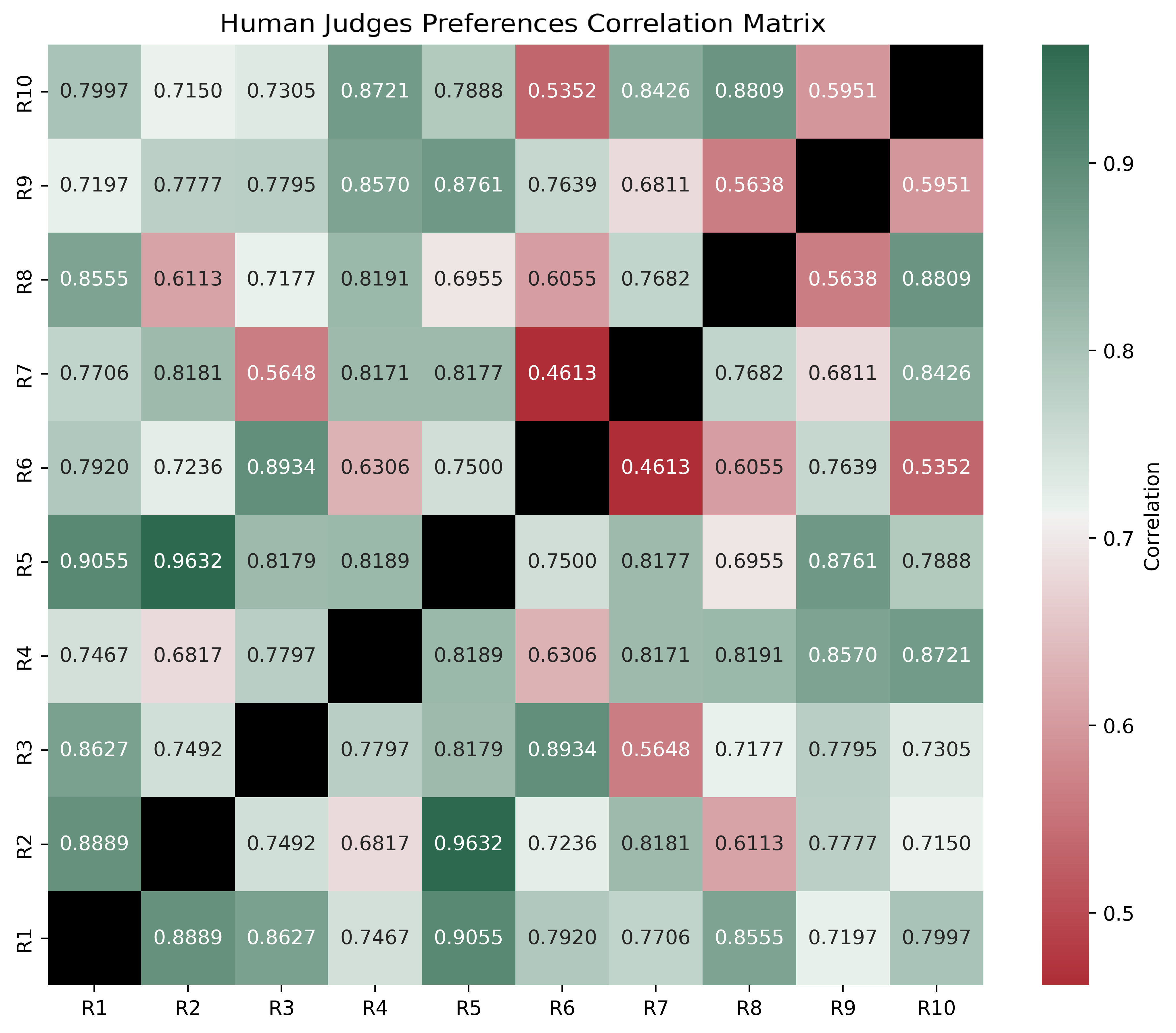}
    \caption{Human Judges Preferences Correlation Matrix}
    \label{fig:vote-correlation}
\end{figure}

Figure~\ref{fig:vote-correlation} presents a correlation matrix of human judges preferences, showcasing the relationships between ratings from different voters (R1 to R10) in the context of model performance comparison. These insights are critical for understanding the consistency and reliability of human evaluations in linguistic model assessments. The matrix reveals both high and low correlations among the human judges' ratings, with values ranging from approximately 0.5 to 0.9. This variation highlights the subjective nature of human evaluation and emphasizes the importance of considering multiple perspectives when assessing model performance.

\begin{figure}[htbp]
    \centering
    \includegraphics[width=0.75\linewidth]{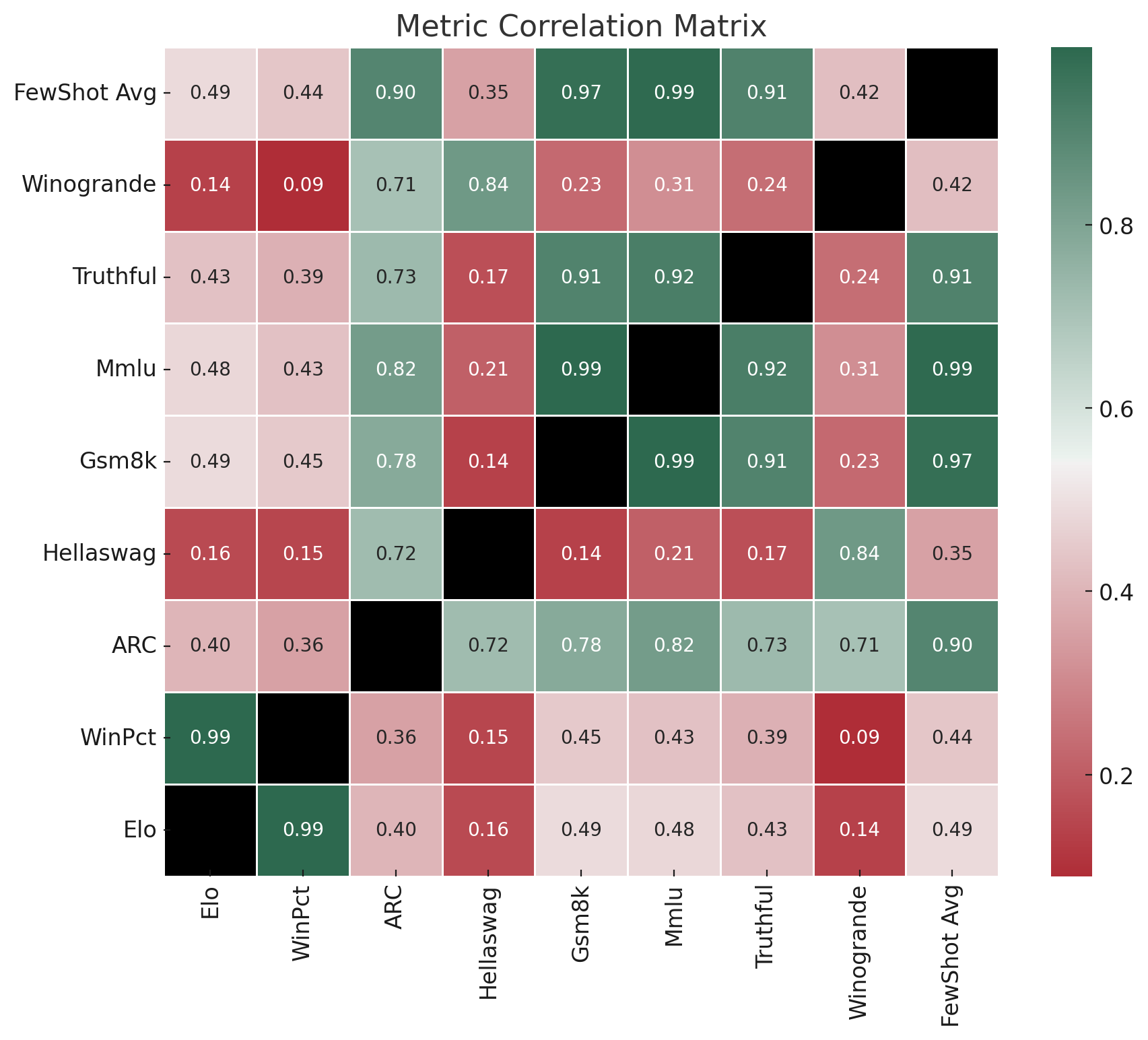}
    \caption{Metric Correlation Matrix}
    \label{fig:Metric-correlation}
\end{figure}

Figure~\ref{fig:Metric-correlation} presents the correlation matrix between different evaluation metrics, that are used to assess the models. Each cell indicates the correlation coefficient between two metrics, with values ranging from 0.0 to 0.9. This matrix helps identify which metrics tend to align closely, indicating that improvements or declines in one metric are often reflected in the other. 

\begin{figure}[htbp]
    \centering
    \includegraphics[width=0.90\linewidth]{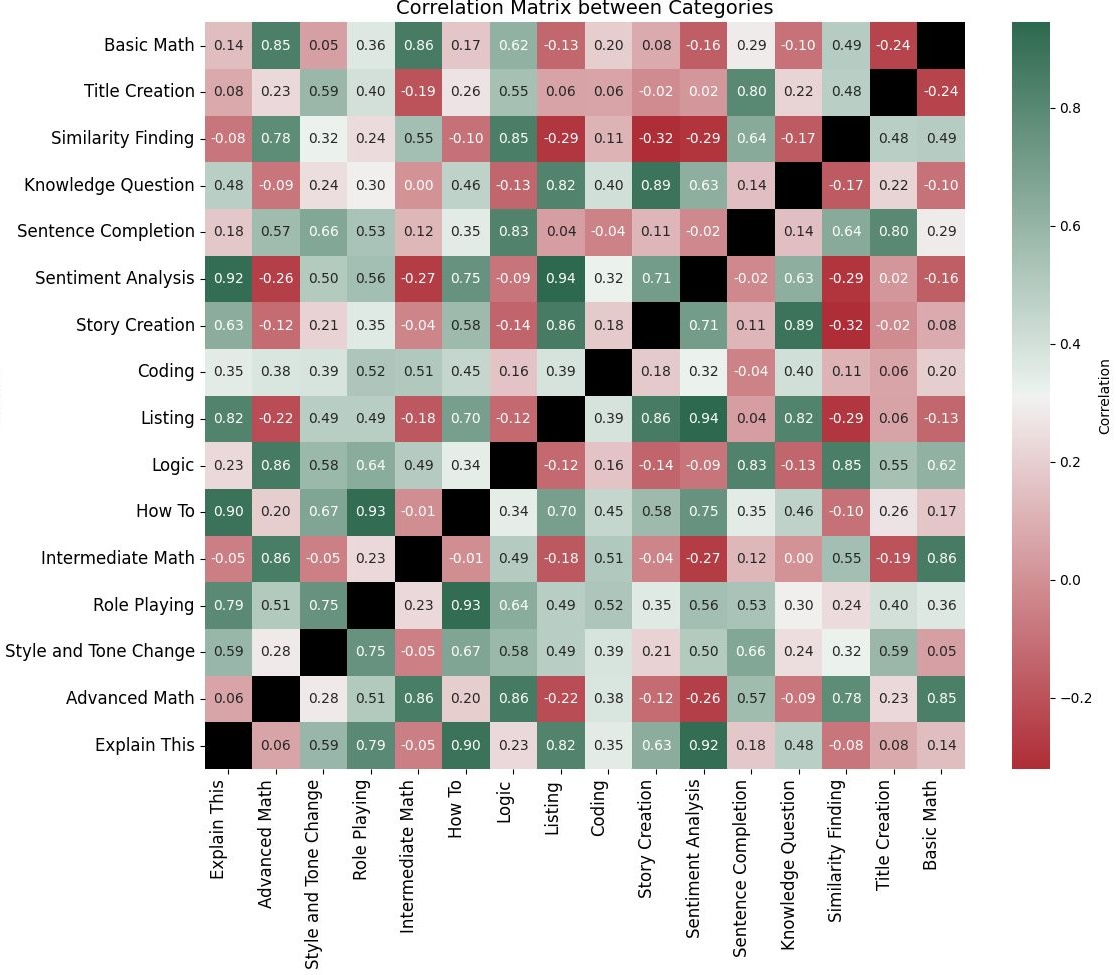}
    \caption{Correlation between categories}
    \label{fig:categorical-correlation}
\end{figure}

Figure~\ref{fig:categorical-correlation} presents the correlation matrix between different evaluation categories, as assessed by voters. Each cell indicates the correlation coefficient between two categories, with values ranging from -0.4 to 0.9. High positive correlations, such as 0.94 between \textit{Sentiment Analysis} and \textit{Sentence Completion}, suggest that performance in one category is strongly associated with performance in the other. This matrix helps identify which categories tend to be evaluated similarly by voters, revealing insights into model strengths and weaknesses across different types of tasks. The correlation matrix between evaluation categories provides insightful observations about the inter-dependencies and distinct relationships among various tasks. Some categories have a low correlation with most of the categories, highlighting the uniqueness of such categories, like \textit{Coding} and \textit{Title Creation}. Analytical categories, such as \textit{Basic Math}, \textit{Intermediate Math}, \textit{Advanced Math}, and \textit{Logic}, are highly correlated illustrating the association and interdependence of the categories.

\section{Conclusion and Future Studies}

This study demonstrates the efficacy of novel adaptation dataset generation, refined corpus selection methodologies, and efficacious training strategies in enhancing the performance of Turkish language models. Our research has revealed that these innovative approaches have led to substantial enhancements in model performance. In particular, optimized corpus selection methodologies and training strategies have enabled Turkish language models to generate more accurate and comprehensive responses. Synthetic datasets have significant potential for languages with limited data sources, such as Turkish. Our study has demonstrated that such datasets play a pivotal role in enhancing the comprehension and responsiveness of language models. Synthetic and translation datasets have been particularly instrumental in addressing Turkish language data gaps and expanding model capabilities.

The findings of the study indicate that enhancements made in small-scale models are reflected in large-scale models in a positive manner. This substantiates the assertion that optimizations made on small models during the model development process provide a robust foundation for the transition to larger and resource-intensive models. Optimizing small models can enhance the overall performance of the model while optimizing cost and resource utilization.


\section*{Acknowledgment}

Research supported with Cloud TPUs from Google's TPU Research Cloud (TRC).

This study was supported by the Scientific and Technological Research Council of Turkey (TUBITAK) Grant No: 124E055.

\bibliographystyle{ieeetr} 
\bibliography{references} 

\end{document}